\documentclass[times,twocolumn,final,authoryear]{elsarticle}

\usepackage{ycviu}
\usepackage{framed,multirow}

\usepackage{amssymb}
\usepackage{latexsym}

\usepackage{url}
\usepackage{xcolor}
\definecolor{newcolor}{rgb}{.8,.349,.1}
\usepackage{amssymb}
\usepackage{amsmath}
\usepackage{booktabs}
\usepackage{hyperref}
\usepackage{subfig}
\usepackage{subfloat}
\usepackage{enumerate}
\usepackage{threeparttable}
\usepackage{color}
\usepackage{makecell}
\usepackage{multirow}

\journal{Computer Vision and Image Understanding}

\begin{document}

\begin{table*}[!t]
% \ifpreprint\else\vspace*{-15pc}\fi

\section*{Research Highlights (Required)}

To create your highlights, please type the highlights against each
\verb+\item+ command. 

\vskip1pc

\fboxsep=6pt
\fbox{
\begin{minipage}{.95\textwidth}
It should be short collection of bullet points that convey the core
findings of the article. It should  include 3 to 5 bullet points
(maximum 85 characters, including spaces, per bullet point.)  
\vskip1pc
\begin{itemize}
\item  We propose an effective single-stage framework for multi-hand 3D reconstruction from a still image. To the best of our knowledge, we are the first to detect and recover the textured hand mesh simultaneously from images in the wild. 
\item Our method supports end-to-end training and has clear advantages over previous similar methods in terms of accuracy and inference speed. Besides, our single-stage framework requires no additional third-party detectors, making it easier to deploy.
\item Both quantitative and qualitative results demonstrate the effectiveness of our proposed framework. Our method achieves the state-of-the-art performance under the weakly-supervised setting, which even outperforms several fully-supervised model-based methods.

\end{itemize}
\vskip1pc
\end{minipage}
}

\end{table*}

\clearpage

\ifpreprint
  \setcounter{page}{1}
\else
  \setcounter{page}{1}
\fi

\begin{frontmatter}

\title{End-to-end Weakly-supervised Single-stage Multiple 3D Hand Mesh Reconstruction from a Single RGB Image}

\author[1]{Jinwei \snm{Ren}} 
\author[1,2]{Jianke \snm{Zhu}\corref{cor1}}
\cortext[cor1]{Corresponding author.}
\ead{jkzhu@zju.edu.cn}
\author[1]{Jialiang \snm{Zhang}}

\address[1]{School of Computer Science and Technology, Zhejiang University, 38 Zheda Road, Hangzhou 310000, China}
\address[2]{Alibaba-Zhejiang University Joint Research Institute of Frontier Technologies, Hangzhou 310000, China}

% \received{1 May 2013}
% \finalform{10 May 2013}
% \accepted{13 May 2013}
% \availableonline{15 May 2013}
% \communicated{S. Sarkar}

\begin{abstract}
In this paper, we consider the challenging task of simultaneously locating and recovering multiple hands from a single 2D image. Previous studies either focus on single hand reconstruction or solve this problem in a multi-stage way. Moreover, the conventional two-stage pipeline firstly detects hand areas, and then estimates 3D hand pose from each cropped patch. To reduce the computational redundancy in preprocessing and feature extraction, for the first time, we propose a concise but efficient single-stage pipeline for multi-hand reconstruction. Specifically, we design a multi-head auto-encoder structure, where each head network shares the same feature map and outputs the hand center, pose and texture, respectively. Besides, we adopt a weakly-supervised scheme to alleviate the burden of expensive 3D real-world data annotations. To this end, we propose a series of losses optimized by a stage-wise training scheme, where a multi-hand dataset with 2D annotations is generated based on the publicly available single hand datasets. In order to further improve the accuracy of the weakly supervised model, we adopt several feature consistency constraints in both single and multiple hand settings. Specifically, the keypoints of each hand estimated from local features should be consistent with the re-projected points predicted from global features. Extensive experiments on public benchmarks including FreiHAND, HO3D, InterHand2.6M and RHD demonstrate that our method outperforms the state-of-the-art model-based methods in both weakly-supervised and fully-supervised manners. The code and models are available at {\url{https://github.com/zijinxuxu/SMHR}}.
\end{abstract}

\begin{keyword}
\MSC 41A05\sep 41A10\sep 65D05\sep 65D17
\KWD End-to-end network \sep 3D Reconstruction \sep Single stage \sep Weakly-supervision \sep Multiple hands
%% MSC codes here, in the form: \MSC code \sep code
%% or \MSC[2008] code \sep code (2000 is the default)
\end{keyword}

\end{frontmatter}

%\linenumbers

%% main text
\section{Introduction}
\label{sec1}
Recently, a surge of research efforts~\citep{zimmermann2019freihand,hampali2020honnotate,chen2021s2hand} have been devoted to 3D hand reconstruction. In contrast to the conventional approaches relying on RGB-D sensor~\citep{yuan2018depth} or multiple view geometry~\citep{simon2017hand}, recovering 3D hand pose and its shape from single color image is more challenging due to the ambiguities in depth and scale. 

% By taking advantage of deep learning techniques, some promising progress on human body~\citep{pavlakos2019expressive} and face reconstruction~\citep{tran2017regressing} has been achieved through neural differentiable rendering. Since hands usually have small size with severe self-occlusions and complex articulations, 3D hand reconstruction is more challenging. To deal with these problems, model-based methods~\citep{boukhayma20193d,zhang2019end,zhang2021hand} make use of 3D parametric model~\citep{romero2017embodied}, and vertex-based  methods~\citep{ge20193d,Kulon2020WeaklySupervisedMH,Chen2021CameraSpaceHM} rely on the graph convolution network (GCN). Except the reconstruction accuracy, inference time, generalization ability and supervision types are also within the scope of this work.  %We follow the former approaches, which can be easily integrated into similar parametric body model in further applications. 

\begin{figure}[t]
\centering
\includegraphics[width=0.98\columnwidth]{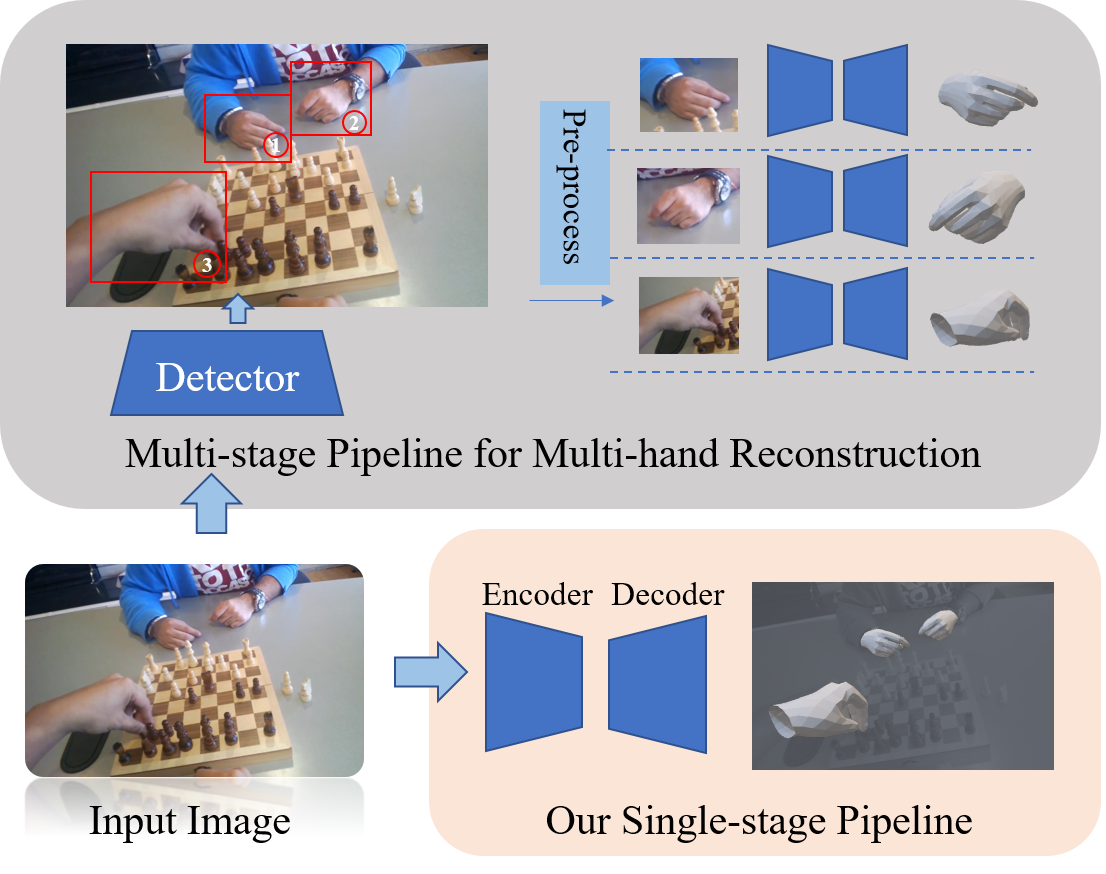} % Reduce the figure size so that it is slightly narrower than the column. Don't use precise values for figure width.This setup will avoid overfull boxes.
\caption{Comparison of conventional multi-stage scheme and our single-stage pipeline. Our method avoids the redundant feature encoding process, which does not rely on the off-the-shelf hand detectors.}
\label{pipelinecompare}
\vspace{-0.2in}
\end{figure}
Most of existing methods mainly focus on the problem of recovering single hand only. However, human naturally uses both of their hands in daily life. In the scene of multi person interaction, such as shaking hands, playing chess, sign language and piano teaching, it is necessary to detect and recover the pose of multiple hands at the same time. Hence, reconstructing multiple hands from a single image is a promising task that has not been extensively studied yet. 
% There are some previous studies try to address this problem. Simon et al.~\citep{simon2017hand} employ multiple view setups while Taylor et al.~\citep{taylor2017articulated} use a high frame-rate depth camera. Mueller et al.~\citep{mueller2019real} present a real-time two hand reconstruction method using single commodity depth camera. Recently, Moon et al.~\citep{moon2020interhand2} propose a 3D dataset for interacting hand pose estimation. These methods either require the extra sensor or assume that there are just two hands in the scene. In order to expand the application scope to a wider range of scenarios, we consider to recover any number of hands from single color image. 
In general, recovering multiple hands in an image is more difficult than reconstructing one hand. A straightforward solution is to decompose it into two separate parts. The hand region is firstly detected by the off-the-shelf object detector, which is further fed into a single hand regressor. However, this two-stage pipeline suffers from the problem of redundant feature extraction. Since it processes each hand instance individually, the overall computation cost grows linearly with the total number of hands in the image. Besides, the hand reconstruction heavily depends on the quality of the detector, which brings the uncertainty and inconvenience in the real-world application. 
% The rich background information is also ignored due to cropping the hand regions. 
In the case of multi-hand scenarios, the relative position in camera space is necessary for scene understanding and interaction analysis. However, the multi-stage model may only deal with the cropped and centered hand and recover the relative pose in the model space.

To address the above limitations, we propose an efficient end-to-end weakly-supervised multiple 3D hand mesh reconstruction approach. Inspired by the single-stage object detection methods, such as CenterNet~\citep{zhou2019objects} we integrate both hand center localization and mesh reconstruction within single network inference. There are some works adopt similar pipeline for multiple human face\citep{Zhang2021WeaklySupervisedM3} and body reconstruction~\citep{Sun2021MonocularOR,Zhang2022Multiperson3P}. However, the hand suffers from severe self-occlusions and complex articulations, which makes it more challenging for estimation. Besides, we need to additionally consider the problem of distinguishing between left and right hands instead of simply treating it as the center of the target. % Differently from those single hand reconstruction methods regressing the hand parameters from the extracted patch, we use the original image and gather the feature vector from the final feature map according to the center map of each hand, which makes it possible to recover multiple hands simultaneously. 
% To this end, we specifically design our network structure to better extract local and global features. 
During the training period, a differentiable hybrid loss upon each hand center is employed to learn the decoupled hand model parameters and position jointly in an end-to-end manner. The comparison of our single-stage pipeline and conventional multi-stage scheme is depicted in Fig.~\ref{pipelinecompare}. 

Besides, it is  extremely difficult to obtain the 3D labels for real-world image, especially in multiple hands scenarios. Methods requiring tedious and time-consuming iterative optimization and numerous fine-grained 3D labels are not friendly enough for the real-world applications. Transformer-based approach~\citep{lin2021endtoend,Lin2021MeshG} and GCN-based methods~\citep{ge20193d,Chen2021CameraSpaceHM} may not be suitable for this scenario, since they often require dense 3D supervision for all vertices and a coarse-to-fine refinement process. 
% Although there are some synthetic multi-hand datasets, the domain shift between different image feature distributions~\citep{cai2018weakly} leads to a large performance drop by training on synthetic dataset or restricted laboratory environment and fine-tuning in real scenarios. 
In contrast, the requirements of model-based method for 3D supervision are not so strict. Thus, we adopt a model-based method trained through purely weakly-supervised fashion to mitigate the dependence on the expensive 3D manual annotated data. 
% In order to narrow the gap with multi-stage and GCN-based methods, we propose a feature aggregation strategy that fully utilizes the 2D cues. The 2D pose estimated from local features serves as a guidance for 3D pose estimated from global features.
We demonstrate our superior performance on single hand dataset FreiHAND~\citep{zimmermann2019freihand} and HO3D~\citep{hampali2020honnotate}, in both weakly-supervised and fully-supervised settings. For the two-handed case, we obtain comparable results with previous fully supervised methods on RHD~\citep{zimmermann2017learning} using only 2D supervision. For the first time, we introduce the multi-hand scenario which contains more than two hands in the same image. Since there is no such dataset publicly available, we generate a multi-hand dataset for training and evaluation. 

Comparing to the previous multi-stage pipeline, our single-stage method benefits from the anchor-free scheme, which can effectively deal with occlusions, as~\citep{Sun2021MonocularOR} does.
% This is because our model is trained by the pixel-wise center map while the previous approaches~\citep{zimmermann2017learning,moon2020interhand2} rely on the box-level detector. 
More importantly, our method has the advantages of fast inference and easy configuration in multi-hand scenarios, as it does not require multiple encodings and gets rid of the limitations of third-party detectors. Besides, the hand texture is important in applications such as augmented reality and virtual reality, which provides more expressive and useful mesh representation. Benefiting from the high scalability of the proposed framework, we can easily extend our model to estimate texture and lighting parameters. 
%It is important to noting that multiple hands within an image share the same intrinsic camera parameters, which is consistent with the actual condition. However, in conventional two-stage pipeline, detecting model and reconstructing model are learned separately, which affects the relative camera pose recovery between the different hands in the same image. Furthermore, the cropped feature may lack necessary information of background, and affect the accuracy of light and texture recovery.

From above all, our main contributions of this work can be summarized as follows:
\begin{enumerate}[(1)]
\item  We propose an effective single-stage framework for multi-hand 3D reconstruction from a still image. To the best of our knowledge, we are the first to detect and recover the textured 3D hand mesh simultaneously from images in the wild. 
\item We design a tight training scheme to optimize the overall framework in a purely weakly-supervised manner. Besides, we propose a multi-hand data augmentation strategy to verify the effectiveness of our method.
\item Our framework supports end-to-end training and has clear advantages over previous similar methods in terms of accuracy and inference speed. Besides, our single-stage framework requires no additional third-party detectors, making it easier to deploy.
\item Both quantitative and qualitative results demonstrate the effectiveness of our proposed framework. Our method achieves the state-of-the-art performance under the weakly-supervised setting, which even outperforms several fully-supervised model-based methods.
\end{enumerate}

\section{Related Work}
\subsection{3D Single Hand Reconstruction.}
Compared to 2D hand pose estimation that only needs to estimate 2D keypoints, 3D hand pose and mesh estimation are more challenging. Specifically, 3D hand pose estimation~\citep{simon2017hand,Krejov2017GuidedOT,spurr2020weakly} only recovers the sparse hand joints while 3D hand mesh reconstruction~\citep{zhang2019end,choi2020pose2mesh,Chen2021CameraSpaceHM} predicts the dense hand mesh with the richer information of hand pose and shape. In this work, we mainly focus on recovering hand mesh from single color image, which is more challenging than the depth image-based methods~\citep{mueller2019real}. Generally, previous studies in this field can be roughly categorized into two groups, including model-based methods for parameter regression and vertex-based approaches for mesh coordinates estimation. 

As for model-based methods, ~\citep{boukhayma20193d} directly regress shape, pose and view parameters of hand model MANO~\citep{romero2017embodied} and supervise with 2D and 3D joints. ~\citep{zhang2019end} adopt a similar framework architecture and add the silhouette information as supervision by a differentiable render~\citep{kato2018neural}. To tackle the problem of lacking 3D annotated real images, ~\citep{zimmermann2019freihand} capture a large single hand dataset with multi-view setup and obtain annotations through an iterative model fitting process. ~\citep{hampali2020honnotate} propose a similar 3D annotation method that focus on hand and object interactions. ~\citep{qian2020html} present the parametric texture model of hands, and combine it with MANO parameters. ~\citep{lv2021handtailor} propose a tailor module to improve the coarsely reconstructed mesh model provided by the hand module. Recently, ~\citep{zhang2021hand} design a cascaded multitask learning backbone to estimate 2D hand pose, mask and mesh simultaneously, which achieves the promising single hand reconstruction performance. 
For vertex-based methods, ~\citep{Moon2020I2L} propose an image-to-lixel prediction network for 3D mesh estimation, which employ the lixel-based 1D heatmap to localize dense mesh vertex position. ~\citep{ge20193d} propose a GCN-based method trained on synthetic dataset and fine-tune on real dataset with the rendered depth map as supervision. Similarly, ~\citep{choi2020pose2mesh} directly regress 3D coordinates using GCN but require 2D human pose as input. ~\citep{Chen2021CameraSpaceHM} extend the GCN-based pipeline with a feature aggregation and 2D-1D registration for pose recovery. Recently, ~\citep{Spurr2021SelfSupervised3H} adopt contrastive learning to take advantage of unlabeled data to improve performance.

\begin{figure*}[ht]
\centering
\includegraphics[width=0.96\textwidth]{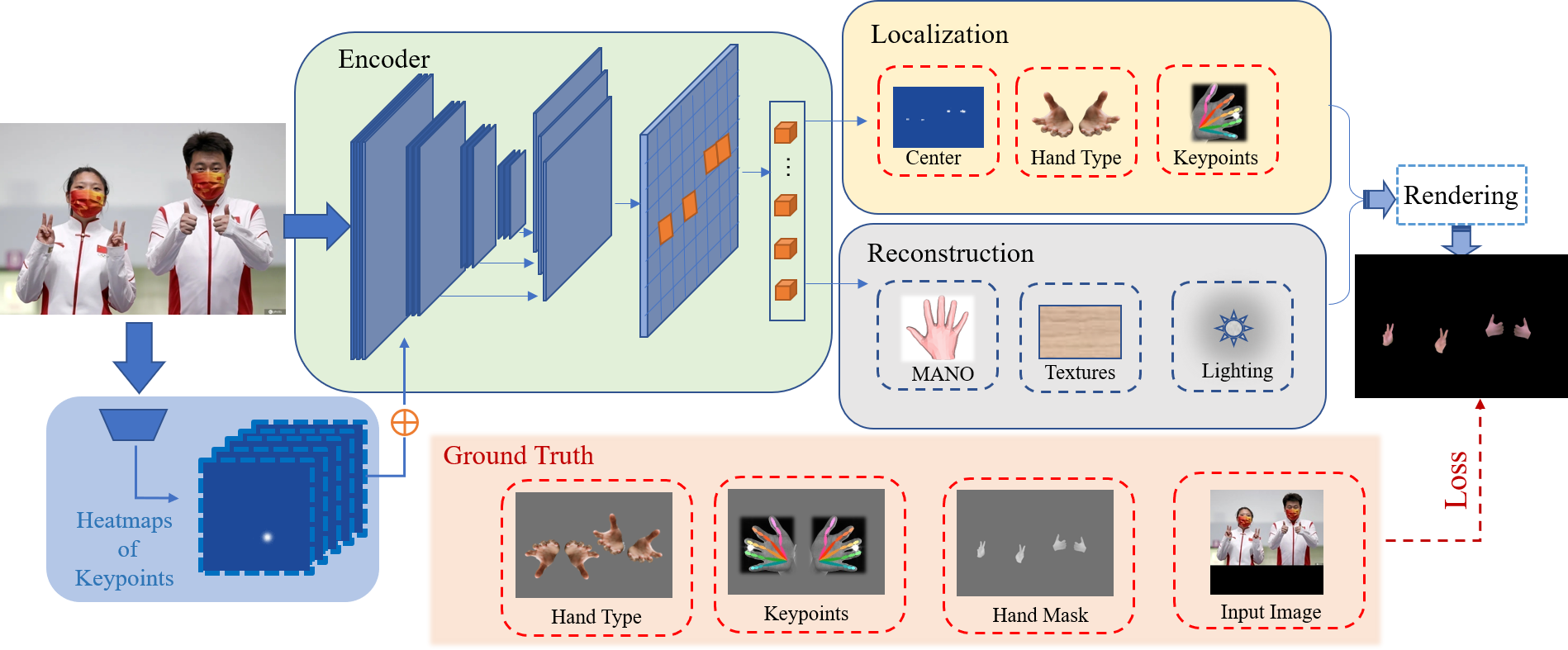} % Reduce the figure size so that it is slightly narrower than the column.
\caption{Overview of the proposed framework. Given an image, we adopt ResNet as backbone to estimate heatmaps of keypoints and concatenate it with the feature map of the first layer. Encoder part  encodes the input features into a center map. Each cell in the map denotes a parametric hand at that position. The localization module is responsible for decoding the hand center, left/right hand type and keypoints and the reconstruction module is responsible for predicting the 3D mesh with MANO, lighting and texture parameters. Each hand mesh is rendered into image space using PyTorch3D~\citep{ravi2020accelerating}. The whole pipeline can be supervised with only 2D labels, which is trained in an end-to-end manner.}
\label{framework}
\vspace{-0.2in}
\end{figure*}

\subsection{3D Multi-hand Estimation}
There are a few existing methods that try to address the 3D multi-hand pose estimation task. ~\citep{mueller2019real} track two hands in real-time using the extra depth sensor. ~\citep{simon2017hand} propose the first 3D markerless hand motion capture system with multi-view setups. ~\citep{zimmermann2017learning} first predict 2D keypoints from color image for both hand and lift them to 3D pose. However, the training images are synthesized from 3D models of humans with the corresponding animations, which are far from realistic. ~\citep{Panteleris2018UsingAS} address this problem in real-world dataset through a three-step pipeline, namely hand area detection, key-points localization and 3D pose estimation. However, several off-the-shelf detectors~\citep{redmon2017yolo9000,simon2017hand} are required in each step. ~\citep{rong2021frankmocap} consider the whole-body capture problem through a standard two-stage pipeline. It firstly detects the body region, and then regresses the human and hand model parameters. Recently, there are some progress~\citep{moon2020interhand2,Li2022InteractingAG} in interacting hand pose estimation. However, bounding boxes of hand area are required for inference. Most of the these methods separately detect and reconstruct multiple hands, which are computational redundant. To this end, we propose a one-stage framework for multi-hand pose estimation. 

\subsection{Weakly-supervised Methods}
% Real-world hand datasets with the accurate 3D annotations are often difficult to obtain, especially for single-view RGB images in the wild. 
In order to get rid of the dependency on massive expensive 3D annotations, some work in recent years tries to estimate 3D hand pose with weak supervision. ~\citep{Neverova2017HandPE} take the depth image as input and fuse it with a novel intermediate representation. ~\citep{cai2018weakly} and ~\citep{Wan2019SelfSupervised3H} adopt a similar pipeline that initializes the network on fully-annotated synthetic data and fine-tunes it on real-world images with depth regularization.
% Kulon et al.~\citep{Kulon2020WeaklySupervisedMH} address the weakly-supervision problem by iteratively fitting hand mesh to image from YouTube videos. Although the mesh annotations are obtained through a weakly-supervised manner, the main network is trained with dense 3D labels using GCN-based method. 
~\citep{spurr2020weakly} introduce a series of biomechanically inspired constraints to guide the hand prediction, including joint skeleton structure, root bone structure and joint angles. ~\citep{chen2021s2hand} employ an off-the-shelf 2D pose detector~\citep{simon2017hand} as a weaker 2D supervision, compared to human annotated 2D keypoints. Differently from the above weakly-supervised methods, our model is designed for multi-hand reconstruction from single image without bounding box. 
% The local keypoint features can then be connected with global center feature for training.  

% \subsection{Multi-hand datasets}

% Since we are the first to explore this field, we can simply treat it as a baseline to verify the efficacy of our framework.

% The author names and affiliations could be formatted in two ways:
% \begin{enumerate}[(1)]
% \item Group the authors per affiliation.
% \item Use footnotes to indicate the affiliations.
% \end{enumerate}
% See the front matter of this document for examples. You are recommended to conform your choice to the journal you are submitting to.

\section{Methodology}
In this section, we present our proposed single-stage multi-hand reconstruction framework. Firstly, we suggest a single-stage pipeline to locate and recover multiple hands simultaneously. Then, we present the localization and reconstruction module, respectively. Finally, a global-local feature consistency loss and multi-hand data augmentation strategy are designed to boost the robustness and accuracy of our proposed approach.

\subsection{Overview}
The overall framework of our method is depicted in Fig.~\ref{framework}, which shares a classical encoder-decoder structure. Given an input image, our model recovers the position as well as 3D pose and shape of each hand in the image. Existing methods~\citep{zimmermann2017learning,moon2020interhand2} address this task by sequentially detecting and reconstructing 3D hand mesh in a multi-stage manner, which incurs extra computational cost on preprocessing hand area and feature extraction. The usage of additional detectors makes such methods not end-to-end. As for our proposed framework, each hand instance is localized and recovered jointly within a single forward pass. To this end, we adopt ResNet-50~\citep{He2016DeepRL} as the backbone of our encoder to extract features, where the parametric hand model MANO~\citep{romero2017embodied} is used as the decoder for hand mesh reconstruction. An optional branch for 2D joint heat-maps estimation is designed to boost the overall performance. Our model predicts the center location, left-right hand type, MANO parameters and rendering parameters, simultaneously.% at the same time.% In the following, we will describe our proposed method in detail.

Our overall training objective function consists of hand localization loss $\mathcal{L}_{loc}$, reconstruction loss $\mathcal{L}_{rec}$ and global-local consistency loss $\mathcal{L}_{con}$ as follows,
\begin{equation}\label{eq1}
\mathcal{L} = \mathcal{L}_{loc} + \mathcal{L}_{rec} + \mathcal{L}_{con}.
\end{equation} 

The localization loss acts as a hand detector in image space. Hand center, keypoints and type are determined by local image feature. The reconstruction loss plays an important role in 3D recovery. Hand pose, shape as well as texture are regressed through global feature sampled from the center map. The consistency loss ensures that the directly estimated and re-projected keypoints are consistent. We describe each module in the following.

\subsection{Multiple Hand Localization}
In this section, we address the problem of hand localization in a 2D image. The input image with resolution of $W \times H$ is divided into $\frac{W}{8} \times \frac{H}{8}$ cells, where each cell represents an individual hand centered in this position. Instead of directly regressing 2D keypoints to estimate hand pose like the conventional method, we predict the center location and left-right hand types to facilitate the subsequent 3D reconstruction.

\noindent \textbf{Deep Hand Encoder}
% Let $I \in \mathbb{R} ^{W \times H \times 3} $ be the input color image containing single or multiple hands, our model outputs a feature map $ Y \in \mathbb{R}^{W/8 \times H/8 \times 256}$ using a stride of 8 for feature aggregation. Then, the feature map is encoded to different semantic vectors through carefully designed full connection layers.
Our backbone follows the network structure design of ResNet-50 for feature extraction. As shown in Fig.~\ref{framework}, we concatenate the feature maps of the last three layers and unify them into the same size of $[b, 256, H/8, W/8]$ via Conv2d and ConvTranspose2d operations. After going through Conv2d, BatchNorm2d and ReLU functions, we get the final feature map. Here $b$ refers to batch size, followed by the number of feature channels and resolution.

As shown in Fig.~\ref{framework}, each cell in the feature map represents an individual hand locating at the corresponding position. The output code vector $\delta$ has all the information to construct a hand, which can be decomposed into center position $\delta_{cp} \in \mathbb{R}^{1}$, left-right hand type $\delta_{lr} \in \mathbb{R}^{2}$, 2D keypoint heat-maps $\delta_{kp} \in \mathbb{R}^{21}$, MANO parameters $\delta_{mano} \in \mathbb{R}^{61}$, texture  coefficient $\delta_{text} \in \mathbb{R}^{778 \times 3}$ and lighting parameters $\delta_{light} \in \mathbb{R}^{27}$. The first three items are used to locate the hand in the 2D image. Moreover, the last three items are used to construct a 3D hand mesh that is rendered into camera space.

% {\bf JK: where is the Appendix? More details about the architecture can be found in Appendix A.}

% \subsection{3D Hand Model Decoder}
\noindent \textbf{Hand Localization} In contrast to the conventional pipeline, we introduce an extra center map to estimate the location for each hand instance. To this end, we employ a heatmap $H \in \mathbb{R}^{\frac{W}{8} \times \frac{H}{8}\times 1}$ to represent the center of each hand, in which each local peak of the probability map indicates a potential hand instance. As discussed in literature~\citep{pfister2015flowing}, the heatmap representation is more robust against noise compared to regressing the pixel coordinate directly. Thus, the hand center coordinates $P_{ct} = \{ p_i \in \mathbb{R}^2 | 1 \leq i \leq k \}$, where $k$ indicates the number of visible hands, is encoded as a Gaussian distribution. The scale of hand is integrated as Gaussian radius. The calculation of radius is referred to~\citep{duan2019centernet}.
% Considering that the center of hand may change according to the gesture, which makes our prediction unstable. We investigate several center definitions and finally choose the mean position of all visible 2D keypoints as ground truth center due to the stable performance across multiple datasets. 
% The detailed analysis can be found in our experiments. 
In the multi-hand setting, hand type has to be considered during training, since the MANO models of left and right hand have different initial position and articulation range. We integrate the left-right hand type into our center map, which is different from face and body reconstruction tasks. 

Given the final feature map, the localization module acts as a decoder, generating the corresponding center, hand type and keypoints predictions. We employ Conv2d to map its output channels into 1(cp), 2(lr), and 21(kp), respectively. The overall loss function of hand localization $\mathcal{L}_{loc}$ consists of three components as follows:

\begin{equation}\label{eq2}
\mathcal{L}_{loc} = \lambda_{cp} \mathcal{L}_{cp} + \lambda_{lr} \mathcal{L}_{lr} + \lambda_{kp} \mathcal{L}_{kp},
\end{equation}
where $\mathcal{L}_{cp}$ refers to the center point localization loss. $\mathcal{L}_{lr}$ denotes the left-right hand type regression loss and $\mathcal{L}_{kp}$ is keypoints detection loss. $\lambda$ is a weighting coefficient to balance the magnitude of different losses. Specifically, $\mathcal{L}_{cp}$ is a modified pixel-wise two-class logistic regression with focal loss~\citep{lin2020focal}. The center of each hand should be categorized as positive class `hand' while the rest area should be treated as negative class `background'. Since there exists imbalance between two kinds of labels, we formulate $\mathcal{L}_c$ like focal loss as below:
\begin{equation}\label{eq3}
\mathcal{L}_{cp} = -\frac{1}{k} \sum_{n=1}^{w \times h} (1 - p_n)^{\gamma} \log(p_n),
\end{equation}
where $k$ is the total number of hands. $p_n \in [0,1]$ is the estimated confidence value for positive class, and $1- p_n$ is the probability for negative class. $w\times h$ is the overall pixel in the center map. $\gamma$ is set to 2 as a hyperparameter to reduce the relative loss for well-classified examples. $\mathcal{L}_{lr}$ also adopts focal loss to solve the problem of imbalance between the positive and negative samples except that we define '0' as left hand and '1' as right hand. $\mathcal{L}_{kp}$ shares the same formulation as $\mathcal{L}_{cp}$ 
with more channels for all keypoints. 

\subsection{Multiple Hand Mesh Reconstruction}
\noindent \textbf{Hand Mesh Representation}
For hand mesh representation, we adopt a model-based method that directly regresses the MANO parameters to shape 3D hand. It has the merit of greatly reducing the search space for poses, which alleviates the difficulty in recovering 3D shape from a single still image. This enables our method to achieve good results with only weak supervision.
% Being compatible with human models such as SMPL, it allows to have better scalability. 
MANO~\citep{romero2017embodied} provides a low-dimensional parametric model to synthesize the hand mesh, which is learned from around 1000 high-resolution 3D hand scans of 31 different persons in a wide variety of hand poses. As in~\citep{romero2017embodied}, we represent with the hand shape $\beta \in \mathbb{R}^{10}$ and pose $\theta \in \mathbb{R}^{51}$ as follows:
\begin{equation}\label{eq4}
M(\beta,\theta)= W(V_P(\beta,\theta),\theta,\mathbb{J}(\beta),\tilde{W}),
\end{equation}

\begin{equation}\label{eq5}
V_P(\beta,\theta) = \bar{V} + \sum_{n=1}^{|\beta|} \beta_n S_n + \sum_{n=1}^{|\theta|} (\theta_n - \bar{\theta} ) P_n,
\end{equation}
where $W$ is the Linear Blend Skinning (LBS) function. $\mathbb{J}$ is a predefined joint regressor, and $\tilde{W}$ is blend weights. Vertices in mesh $V_P \in \mathbb{R}^{778 \times 3}$ are calculated according to shape and pose displacements of template hand mesh $\bar{V}\in \mathbb{R}^{778 \times 3}$. $S_n$ are the principal components in a low-dimensional shape basis and $P_n$ are pose blend shapes controlling vertices offset. $\bar{\theta}$ is the mean pose. With the joint regressor $\mathbb{J}$, we can further calculate the accurate 3D joints $J \in \mathbb{R}^{21 \times 3}$ from the position of the vertices, corresponding to 21 keypoints in image space. It is worthy of mentioning that both global hand rotation and translation are encoded in MANO parameters, which are used in the camera projection. In order to narrow the gap between reconstruction results and real-world application, we estimate the texture parameters for each hand. As in~\citep{chen2021s2hand}, we directly regress per-vertex RGB value $\delta_{text} \in \mathbb{R}^{778 \times 3}$ of each hand mesh. Furthermore, we employ spherical harmonics~\citep{Ramamoorthi2001AnER} to approximate illumination changes of the scene. The estimated lighting vector $L \in \mathbb{R}^{27}$ works well in several datasets with various illumination conditions.

\noindent \textbf{Camera Model}
Based on the above hand mesh representation, we are able to estimate 3D hand in hand-relative coordinates. Projecting each hand into camera-relative coordinate and image coordinate system is essential for the applications. Instead of assigning different camera parameters to each individual hand like conventional multi-stage pipeline, a unified and consistent camera model is more convenient and reasonable. Therefore, we use the same intrinsic matrix $K$ for perspective projection in each dataset during training. By predicting the individual rotation and translation matrix $T = \{[R_i|t_i] \in \mathbb{R}^{3 \times 4} | 1 \leq i \leq k \}$, all hand meshes are transformed into a unified camera coordinate system as follows:
$$P_i^c = KT_iP_i^w$$
\begin{equation}\label{eq6}
K = 
\begin{bmatrix}
f_x & 0 & c_x \\
0 & f_y & c_y \\
0  & 0 & 1 \\
\end{bmatrix},
T_i = 
\begin{bmatrix}
r_{11} & r_{12} & r_{13} & t_1 \\
r_{21} & r_{22} & r_{23} & t_2 \\
r_{31} & r_{32} & r_{33} & t_3 \\
\end{bmatrix},
\end{equation}
where $f_x$, $f_y$ are the focal length fixed as 512 in multi-hand setting, $c_x = W/2$, $c_y=H/2$ are the projection center of the image. Global rotation and translation matrices are estimated in $\delta_{mano}$ together with other joint rotations on Rodrigues vector representation. $P^c \in \mathbb{R}^3$ is the hand mesh in camera coordinate system, and $P^w \in \mathbb{R}^3$ in world coordinate system. $P^w$ is further expanded to a homogeneous coordinate system to calculate the matrix projection. Comparing to the conventional multi-stage methods, our approach enjoys the benefits of coherent environment light and projection model, while the cropped hand patch may lose some precision of texture and scale information. Besides, the important relative position of each hand can be easily recovered in the proposed pipeline without requiring the intrinsic matrix of each hand.

\noindent \textbf{Hand Mesh Reconstruction}
As shown in Fig.~\ref{framework}, the reconstruction module also acts as a decoder, which generates the corresponding MANO, texture and light parameters. We employ Conv2d to map its output channels to parameter scales. This is the same as the localization module above. Based on the estimated MANO parameters and camera model, we are able to render hand mesh into camera space. Given an input image, our model first estimates the center map which represents all visible hands in 2D space. We use max pooling operation to find the local maximums and gather hand parameters according to these indexes. 3D hand meshes $T_P \in \mathbb{R}^{778 \times 3}$ and joints $J \in \mathbb{R}^{21 \times 3}$ are determined by $\delta_{mano} \in \mathbb{R}^{61}$, which are converted to the camera coordinate system through the estimated global rotation and translation terms. Further, we adopt 2D keypoints re-projection loss and photometric loss to learn our parameters as below:
\begin{equation}\label{eq7}
\mathcal{L}_{rec} = \lambda_{rep} \mathcal{L}_{rep} + \lambda_{pho} \mathcal{L}_{pho} + \lambda_{reg} \mathcal{L}_{reg},
\end{equation}
where $\mathcal{L}_{rep}$ refers to the re-projection loss. $\mathcal{L}_{pho}$ is the photometric loss, and $\mathcal{L}_{reg}$ represents the regularization loss. Specifically, $\mathcal{L}_{rep}$ is the sparse 2D keypoints re-projection error that minimizes the distance between 2D projection from its corresponding 3D joints and the labelled 2D ground truth.
\begin{equation}\label{eq8}
\begin{split}
\mathcal{L}_{rep} = \frac{1}{k\times J} \sum_{n=1}^{k} \sum_{j=1}^{J} ||\phi_{n,j} - \phi_{n,j}^*||_2  \\
 + \frac{1}{k\times E} \sum_{n=1}^{k}\sum_{e=1}^{E}||e_{n,e} - e_{n,e}^*||_2
\end{split}.
\end{equation}
In $\mathcal{L}_{rep}$, $J$ is the total number of 2D keypoints, and $E$ is the total number of normalized edge vectors constructed based on each two adjacent keypoints. They correspond to 21 joints and 20 movable bones in the physical sense. $\phi_{n,j}$ refers to the $n^{th}$ hand and $j^{th}$ keypoint projected on image. $e_{n,e}$ is the $n^{th}$ hand and $e^{th}$ bone. Similarly, $*$ indicates the ground truth. We use the length of the first section in the middle finger to unify the errors for different hand scales.

\begin{equation}\label{eq9}
\begin{split}
\mathcal{L}_{pho} = \frac {\sum_{n=1}^{W \times H} M_{n}||I_{n} - I_{n}^*||_2 } {\sum_{n=1}^{W \times H}M_{n}}.
\end{split}
\end{equation}
$\mathcal{L}_{pho}$ is the photometric error between the input and rendered images. Hand mask is used to exclude the influence of irrelevant background pixels. $I$ and $I^*$ are the rendered and input images, respectively. $M$ is a binary mask with the same size of input image, which is determined by Hadamard product between the rendered silhouette  and ground truth skin mask. Since we ignore the pixels in background area, $M_n$ in such positions is just set to zero. 3D textured hand mesh is constructed with $\delta_{mano}$, $\delta_{text}$ and $\delta_{light}$ and the rendering is implemented through PyTorch3D~\citep{ravi2020accelerating}. 

$\mathcal{L}_{reg}$ is a pose and shape regularization term to penalize the implausible 3D joint rotation and shape. We define an interval $[\theta_{min},\theta_{max}]$ of valid rotation range for each joint angle, since the angle within the range should not be regularized. Shape parameters are encouraged to be close to the mean value. In this paper, we define the regularization loss as below:
\begin{equation}\label{eq10}
\begin{split}
\mathcal{L}_{reg} &=  w_{pose}||\delta_{pose}||_1+w_{shape} ||\delta_{shape}||_2, \\
\end{split}
\end{equation}
where $\delta_{pose}$ is the pose error that penalizes $\theta$ exceeding the predefined threshold, and $\delta_{shape}$ is the shape error pulling $\beta$ to be close to mean shape.

\noindent \textbf{Global-local Feature Consistency}
To further improve the performance, we consider combining the global feature and local feature together. Specifically, the 2D keypoints directly estimated from local features and re-projected points from 3D joints estimated from global features should be equal. Interestingly, our center map plays an important role in top-down estimation while our 2D keypoints heatmap is essential to bottom-up estimation.
\begin{equation}\label{eq11}
\begin{split}
\mathcal{L}_{con} = \frac{1}{k\times J} \sum_{n=1}^{k} \sum_{j=1}^{J} ||\phi_{n,j}^{kp} - \phi_{n,j}^{rep}||_2  .
\end{split}
\end{equation}
Similar ideas have been addressed in single-hand methods as intermediate supervision. In multi-hand issue, however, we have to consider the joint assignment task. Keypoints belong to the same hand should be close to each other while different hands should fall apart from each other. We use L2 normal to constrain the above assumptions.

\subsection{Weakly-supervised Data Augmentation}
% Since most of previous studies focus on single-hand recovery task, dataset for 3D multi-hand pose estimation is rare. FreiHAND~\citep{zimmermann2019freihand} and HO3D~\citep{hampali2020honnotate} are designed for single-hand estimation. RHD~\citep{zimmermann2017learning} is a stereo hand dataset while there is always a big gap between the synthetic data and real image. Tzionas et al.~\citep{tzionas2016capturing} focus on hands in action~(hand-hand interaction and hand-object interaction) with RGBD setup, where only 2D ground-truth joints are provided. The dataset is relatively small in size. Simon et al.~\citep{simon2017hand} generate hand label through a multi-view bootstrapping method for images captured from Panoptic Studio dataset. Unfortunately, the hand part is rather small compared to image resolution, which leads to the inaccurate annotations. Recently, InterHand2.6M~\citep{moon2020interhand2} provides a large-scale real-captured hand dataset with 3D annotation. However, the dataset with only two hands still restricts the algorithm to the limited scenarios. Bambach et al.~\citep{Bambach_2015_ICCV} provides the first-person interactions between two people, where up to four hands may appear at the same time. However, only mask labels of visible hands are provided in this dataset. Therefore, we extend the previous hand datasets to a more general form and propose our novel single-stage multi-hand recovering framework based on our synthesized dataset. 

To address the problem of lacking annotated training data with multiple hands, we synthesize a large multiscale multi-hand dataset, whose hand samples are selected from the existing single hand datasets, as shown in Fig.~\ref{visualmulti}. This dataset is mainly used to verify the correctness of our method, and can also be used as pre-training data for other multi-hand methods. Specifically, we crop up to 10 hand samples and resize them to paste on an empty image $I \in \mathbb{R}^{512 \times 512}$. To generate more photo-realistic image, we firstly paste one whole image containing single hand and background, which is resized to $I$ and padded with background pixels. Then, we crop other hand samples according to its bounding box and randomly re-scale them into the size between 96 pixels to 320 pixels. For simplicity, we crop each hand in the original image into a square. The irrelevant background pixels are removed by hand masks. To place the hand samples, we use the greedy strategy to set them layer by layer from the lower right corner to the upper left corner in the image. The size of the next sample is randomly generated according to the remaining available space until the rest available space is less than the predefined minimum sample size. As for ground truth construction, we retain the index of each hand sample, 2D keypoints, center point, bounding box and mask, which are generated from the original data labels by affine transformation. Besides, we randomly flip the original hand patch horizontally to construct a left-hand image for hand type learning, since all images in FreiHAND and HO3D only have right hands.

\begin{table*}[htp]
\caption{Comparison with state-of-the-art model-based} methods on the FreiHAND dataset. Bold represents the best result and underlined represents the second-best result.
\centering
\resizebox{.95\textwidth}{!}{
\begin{tabular}{ccccccccc}
\toprule
    Methods & supervision & camera intrinsic & MPJPE$\downarrow$ & $AUC_J$$\uparrow$ & MPVPE$\downarrow$  & $AUC_V$$\uparrow$ & $F_5$$\uparrow$ & $F_{15}$$\uparrow$\\
\midrule
    Boukhayma et al.~\citep{boukhayma20193d} & 3D & Yes &  3.50 & 0.351 & 1.32 & 0.738 &0.427 & 0.895\\
    ObMan~\citep{hasson2019learning} & 3D & Yes &  1.33 & 0.737 & 1.33 & 0.736 &0.429 & 0.907\\    
    ManoCNN~\citep{zimmermann2019freihand} & 3D & Yes  & 1.10 & 0.783 & 1.09 & 0.783 &0.516 & 0.934 \\
    ManoFit~\citep{zimmermann2019freihand} & 3D & Yes  & {1.37} & 0.730 & {1.37} & 0.729 &0.439 & 0.892 \\    
    % Biomechanical~\citep{spurr2020weakly} & 3D & Yes  & 0.90 & 0.820 & - & - & - & - \\    
    HTML~\citep{qian2020html} & 3D & Yes &  1.11 & 0.781 & 1.10 & 0.781 & 0.508 & 0.930 \\
    HIU(single)~\citep{zhang2021hand} & 3D & Yes &  0.89 & 0.824 & 0.92 & 0.819 & 0.571 & 0.961 \\    
    HIU(cascaded)~\citep{zhang2021hand} & 3D & Yes &  \textbf{0.71} & \textbf{0.860} & \textbf{0.73} & \textbf{0.856} & \textbf{0.699} & \textbf{0.974} \\    
    HandTailor~\citep{lv2021handtailor} & 3D & Yes & 0.82 & - & 0.87 & - & - & - \\  
    Ours & 3D & No & \underline{0.80} & \underline{0.840} & \underline{0.81} & \underline{0.839} & \underline{0.649} & \underline{0.966} \\  
    
\midrule
    Biomechanical~\citep{spurr2020weakly} & 2D & Yes  & 1.13 & 0.780 & - & - & - & - \\
    S2HAND~\citep{chen2021s2hand} & 2D & Yes  & 1.18 & 0.766 & 1.19 & 0.765 & 0.48 & 0.92 \\
    Ours & 2D & No & \textbf{1.07} & \textbf{0.788} & \textbf{1.10} & \textbf{0.782} & \textbf{0.500} & \textbf{0.937} \\
\bottomrule
\end{tabular}
\label{table1}}
\end{table*}

\begin{table*}[ht]
\caption{Comparison with previous model-based methods on HO-3D evaluation dataset.}
\centering
%\resizebox{.95\columnwidth}{!}{
\begin{tabular}{cccccccc}
\toprule
    Methods & supervision & MPJPE$\downarrow$ & $AUC_J$$\uparrow$ & MPVPE$\downarrow$  & $AUC_V$$\uparrow$ & $F_5$$\uparrow$ & $F_{15}$$\uparrow$\\
\midrule		
    HO3D~\citep{hampali2020honnotate} &3D  &1.07 & 0.788 & 1.06 & 0.790 & 0.51 & 0.94 \\
    ObMan~\citep{hasson2019learning} & 3D & - & - & 1.10 & 0.780 & 0.46 & 0.93 \\
    Photometric~\citep{hasson2020leveraging} & 3D & 1.11 & 0.773 & 1.14 & 0.773 & 0.43 & 0.93 \\
    Ours & 3D & \textbf{1.01} & \textbf{0.799} & \textbf{0.97} & \textbf{0.805} & \textbf{0.524} & \textbf{0.953} \\    
\midrule
    PeCLR~\citep{Spurr2021SelfSupervised3H} & 2.5D & 1.09 & 0.78 & - &- & - & - \\
    S2HAND~\citep{chen2021s2hand} & 2D & 1.14 & 0.773 & 1.12 &0.777 & 0.45 & 0.93 \\
    Ours & 2D & \textbf{1.03} & \textbf{0.794} & \textbf{1.01} & \textbf{0.797} & \textbf{0.502} & \textbf{0.951} \\
\bottomrule
\end{tabular}
\label{table2}
\end{table*}

\begin{table}[!ht]
\caption{EPE comparison on RHD dataset. GT S and GT H denote ground truth scale and hand type~(left/right), respectively. y means ground truth used during inference while n means not used.}
\centering
\resizebox{.95\columnwidth}{!}{
\begin{tabular}{cccc}
\toprule
    Methods & GT S & GT H & EPE$\downarrow$ \\
\midrule
    RHD~\citep{zimmermann2017learning} & y & y & 3.04  \\
    yang2019disentangling~\citep{yang2019disentangling} & y & y &  1.99 \\
    spurr2018cross~\citep{spurr2018cross} & y & y & \textbf{1.97} \\
\midrule
    spurr2018cross~\citep{spurr2018cross} & n & n & 2.25 \\  
    InterNet~\citep{moon2020interhand2} & n & n & 2.08 \\      
    Ours & n & n & \textbf{2.07} \\ 
\bottomrule
\end{tabular}}

\label{table3}
\end{table}

\section{Experiment}
In this section, we thoroughly evaluate our proposed framework. Firstly, we present the implementation details for experimental setup. Then, the comprehensive experiments are conducted in order to compare with the state-of-the-art methods, including single-hand setting, two-hand setting and multi-hand setting. Finally, we give an ablation study to examine the effect of each individual module and give the potential direction for further improvement.

\subsection{Implementation Details}
The proposed framework is implemented with PyTorch~\citep{Paszke2019PyTorchAI}. Our method can simultaneously train detection and reconstruction modules in end-to-end. To facilitate exploring the impact of different modules on model performance, we also design a staged training strategy. In the first stage, we reduce the model task to single hand reconstruction and only update the reconstruction parameters. In the second stage, we perform joint training of detection and reconstruction based on the model obtained in the previous stage. 
% For performance comparison, please refer to the RHD dataset part of the two-hand reconstruction experiment.
% For the sake of experimental completeness, we compare the performance and time between {\bf ??? multi-stage training} and single-stage joint training in the ablation study. 
In our training process, the batch size is set to 256, and the initial learning rate is $10^{-3}$. We decrease our learning rate by 10 at the epoch of 30, 90 and 120. We train our model with four RTX2080Ti GPUs, which takes around a day to train 70K iterations on FreiHAND dataset. The input images are resized into $224\times224$ for single-hand estimation task and $512\times512$ for multiple hand recovering task. The typical data augmentation methods, including random scaling, translation, rotation and color jittering, are performed in both single and multiple hand settings.  

\subsection{Datasets and Evaluation Metrics}

\noindent \textbf{FreiHAND}~\citep{zimmermann2019freihand} is a large-scale single hand dataset with 3D labels on hand joints and MANO parameters. The evaluation set contains 3960 samples without ground truth annotations. Researchers need to submit their predictions to the online server for evaluation. The training set contains 32,560 samples of real human hands captured with green screen background, and each sample is augmented with different synthetic backgrounds. The whole dataset contains 130,240 images with various hand poses and augmented background using different post-processing options.
% Four training samples in each group share the same 3D label, segmentation mask, camera intrinsic matrix, 3D keypoint annotation and their corresponding MANO parameters. All hands are centered in the original images with the size of 224$\times$224.

\noindent \textbf{HO-3Dv2}~\citep{hampali2020honnotate} is a single hand dataset with 3D annotations similar to FreiHAND, which focuses on hand-object pose estimation. It contains 68 sequences captured with 10 different persons manipulating 10 different objects. The training set has 66,034 images (from 55 sequences) and the evaluation set contains 11,524 images (from 13 sequences). 
% As the annotation for testing set is unavailable, we submit our results to online server for evaluation. Hands in this dataset are not centered and suffer from sever partial occlusions, which brings the extra challenges for hand recovering. 

% \textbf{OneHand10K,wang2019mask} is a small hand dataset with 2D annotations. It contains 10K RGB images with single hand for training. Images in the dataset are collected from internet while hand segmentation and 2D keypoints are manually annotated. 

\noindent \textbf{RHD}~\citep{zimmermann2017learning} is a large-scale synthetic dataset with 3D annotations for single and interacting hand poses. It is created from 3D models of humans animations using commercial software. The dataset is built upon 20 different characters performing 39 actions, providing 41,258 images for training and 2728 images for evaluation. 

\noindent \textbf{InterHand2.6M}~\citep{moon2020interhand2} is a large-scale real-world dataset with 3D annotations for interacting hand poses. It contains 1,179,648 interacting hand frames and 1,410,699 single hand frames.
% The annotation contains bounding box, 3D joints, camera intrinsic matrix and hand type information.

\noindent \textbf{Evaluation Metrics} As in~\citep{zimmermann2019freihand,hampali2020honnotate}, we evaluate our approach by calculating the errors of 3D joints and 3D vertices. We compute the mean per joint position error (MPJPE) and mean per vertex position error (MPVPE) between the prediction and ground truth in cm for 3D joints and 3D mesh evaluation, respectively. All results on FreiHAND and HO-3D are submitted to online server that aligned automatically based on Procrustes analysis~\citep{Gower1975GeneralizedPA} for fair comparison. We also calculate the area under curve ($AUC_J$ for joints and $AUC_V$ for vertices) of the percentage of correct keypoints (PCK) curve in an interval from 0 cm to 5 cm with the 100 equally spaced thresholds. Besides, end point error (EPE) is used in two-hand setting, which is defined as a mean Euclidean distance (cm) between the predicted  3D hand pose and ground-truth after root joint alignment. As for 2D keypoint evaluation, we calculate MPJPE using 2D distance in $pixel$.

\begin{figure*}[!ht]
    \centering
    \subfloat[]{\includegraphics[width=1.0\columnwidth]{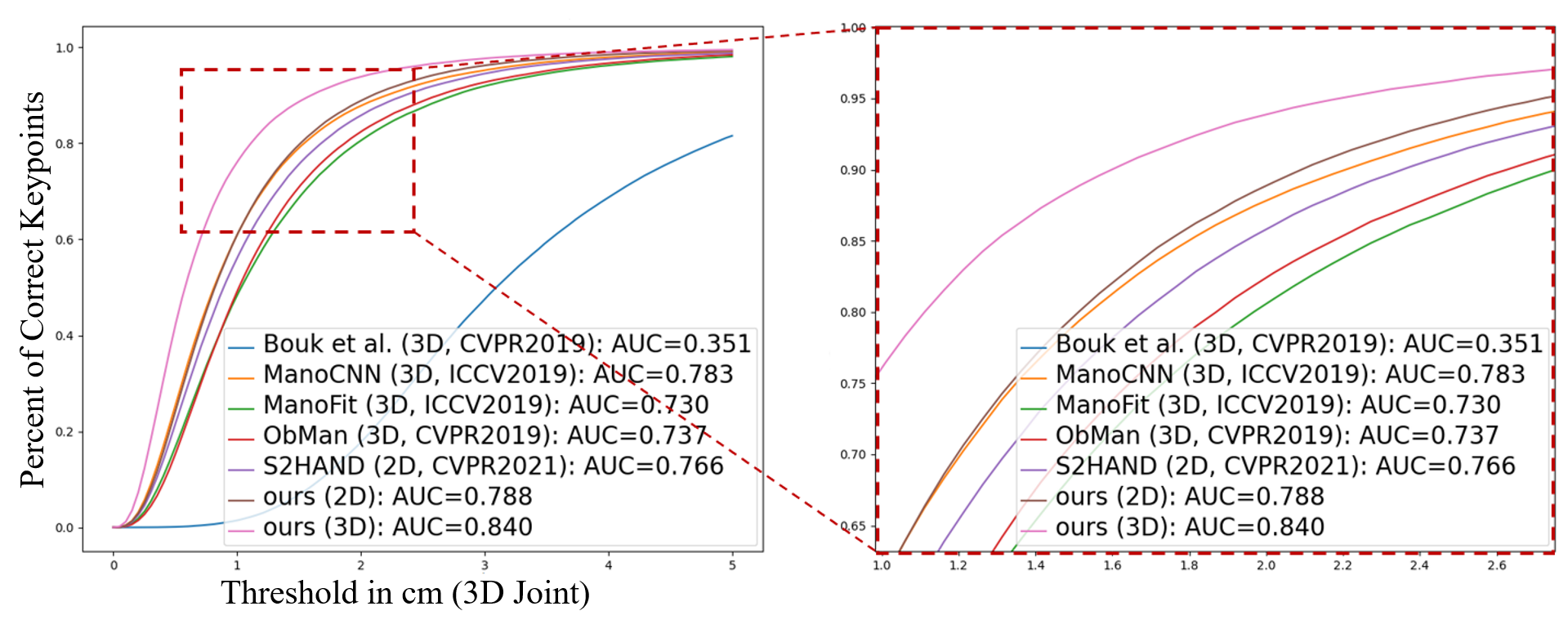}%
    \label{fig_xyz_pck}}
    \hfil
    \subfloat[]{\includegraphics[width=1.0\columnwidth]{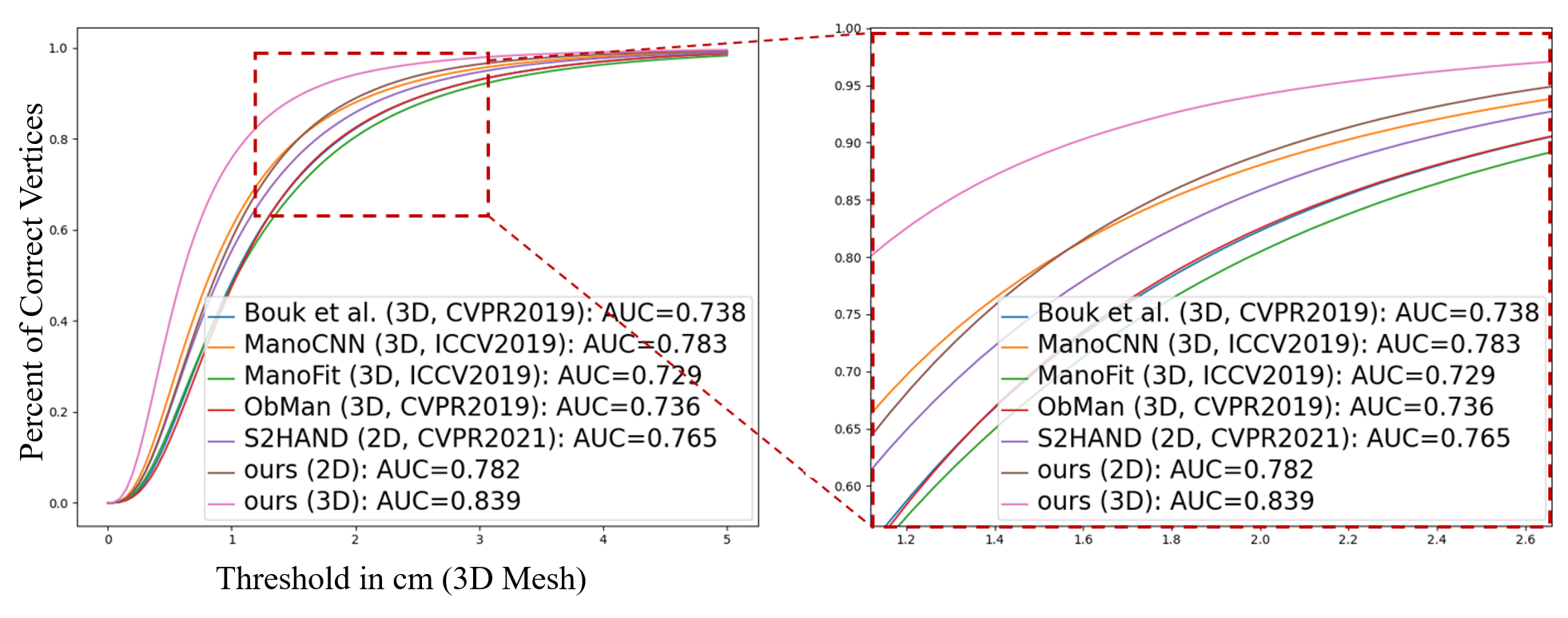}%
    \label{fig_mesh_pck}}
    \caption{3D PCK comparison with state-of-the-art methods on FreiHAND dataset. The left two figures show the result of $AUC_J$ and locally enlarged details. The right two figures show the result of $AUC_V$ as well as locally enlarged details.}
    \label{PCK}
\vspace{-0.1in}
\end{figure*}

\begin{figure*}[ht]
    \centering
    \includegraphics[width=0.98\textwidth]{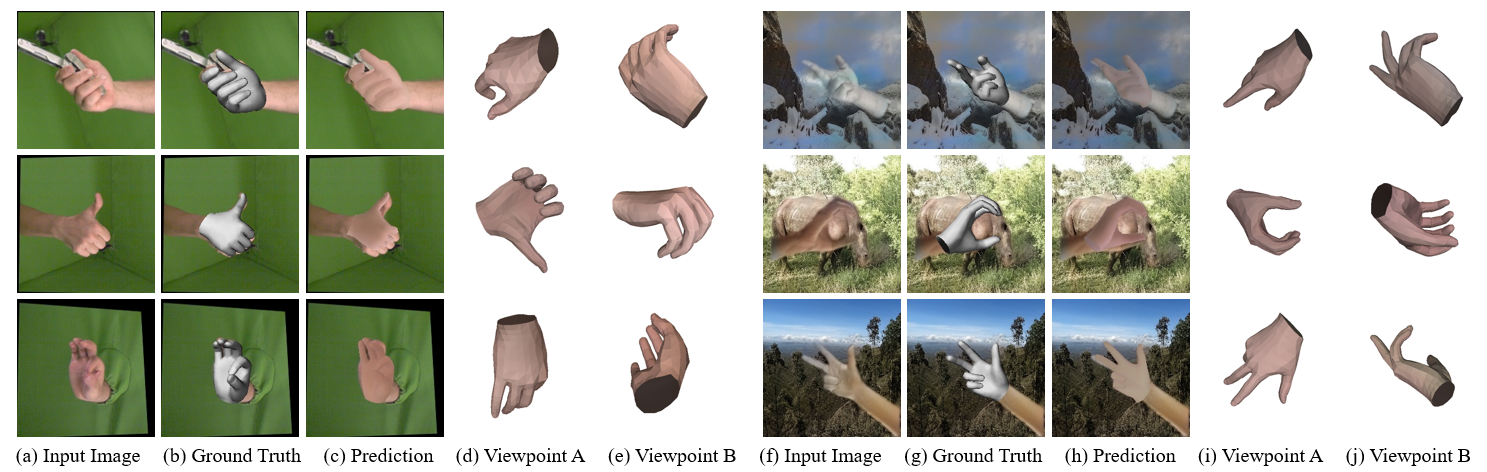} % Reduce the figure size so that it is slightly narrower than the column.
    \caption{Qualitative mesh results on dataset FreiHAND. Our model is trained only with 2D supervision, which do not require the real camera intrinsic parameters. Textures are learned through a self-supervised manner, which makes our outputs more vivid.}
    \label{FreiHANDmesh}
\vspace{-0.2in}
\end{figure*}

\subsection{Comparisons with State-of-the-art Methods}
\noindent \textbf{Single-hand Reconstruction Experiments} We firstly compare our method against the recent state-of-the-art methods in fully-supervised~\citep{boukhayma20193d,zimmermann2019freihand,hasson2019learning,qian2020html,hasson2020leveraging,hampali2020honnotate,zhang2021hand,Spurr2021SelfSupervised3H} and weakly-supervised manner~\citep{spurr2020weakly,chen2021s2hand}. For fair comparison, we mainly focus on the more relevant model-based methods and some excellent We choose FreiHAND and HO-3D as our validation datasets, since they are the latest and mostly used single hand datasets. Evaluations are conducted through submitting our estimated results to their online judging system\footnote{https://competitions.codalab.org/competitions/21238}, \footnote{https://competitions.codalab.org/competitions/22485}. 

Table~\ref{table1} shows the evaluation results on FreiHAND. It can be clearly seen that our presented method outperforms other weakly-supervised methods on all evaluation metrics, which achieves the comparable accuracy against many recent fully-supervised methods. It obtains 1.07 cm MPJPE with 0.787 $AUC_J$ and 1.10 cm MPVPE with 0.782 $AUC_V$. To further explore the potential of our model, we trained it with full supervision when 3D labels are available. Our model also achieves state-of-the-art performance among model-based methods. Note that the HIU (cascaded) in Table~\ref{table1} employs up to 8 cascaded networks following a coarse-to-fine design while our model only uses a single encoder. For fair comparison, we also report the results of HIU (single) from its ablation experiments, using the same single encoder-decoder structure as ours. Our method performs better while having the same model complexity. This reflects the potential of our framework. Fig.~\ref{PCK} plots our 3D PCK of joints and vertices with other methods under different error thresholds. Our fully-supervised model outperforms other methods to a large margin, while our weakly-supervised model achieves the comparable performance against ManoCNN~\citep{zimmermann2019freihand}. 
% Note that Biomechanical~\citep{spurr2020weakly} additionally uses 3D labels of synthetic training data and only predicts the sparse joint poses. 
In the close-up figure, it can be found that our weakly-supervised model is not as good as ManoCNN under the small error thresholds while our method performs better under the large thresholds. This is because it is hard for our method to learn the detailed 3D pose with only 2D label. However, we can achieve generally consistent and fine-grained accuracy. To evaluate 2D pixel error, we randomly select 10\% of the training set for validation, since no ground-truth 2D keypoints available on evaluation set. We train our model with the rest samples of the training set, which obtains 6.64 pixel error/1.29 cm joint absolute error under the input size of $224\times 224$ using 2D supervision. Moreover, we obtain 5.88 pixel error/0.65 cm joint absolute error with 3D supervision. The close pixel error further demonstrates that our presented method can fully make use of 2D supervision to learn the accurate 3D poses, while 3D supervision can disambiguate the perspective projection to further improve performance. Visual results on validation set are depicted in Fig.~\ref{FreiHANDmesh}, which include the input image, ground-truth mesh overlaid on input image, predicted mesh overlaid on input image and textured mesh in two viewpoints. By taking advantage of the photometric loss, our model is able to learn the lighting and texture parameters from input image through a self-supervised manner, which produces more vivid hand mesh. As shown in Fig.~\ref{FreiHANDmesh}, two sets of images from different viewpoints were rendered using open-source system MeshLab~\citep{Cignoni2008MeshLabAO} without lighting. %so there is a little difference in texture from the original image. 
  
\begin{figure}[ht]
    \centering
    \includegraphics[width=0.95\columnwidth]{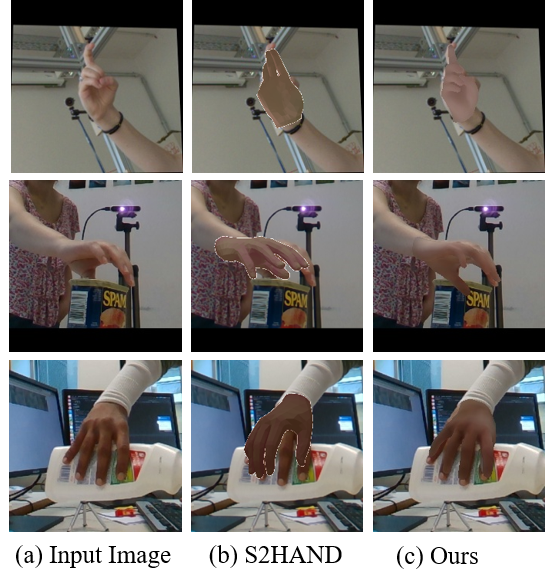} % Reduce the figure size so that it is slightly narrower than the column.
    \vspace{-0.16in}
    \caption{Visualization comparison with the state-of-the-art method on single hand datasets. Our method achieves significant advantages in both hand pose accuracy and texture realism.}
    \label{Visualcompare}
\end{figure}

HO3D is a more challenging dataset for the hand-object interaction containing motion sequences. Hands are usually occluded by the object or partly outside the screen, which makes it even more challenging for our presented method to estimate the hand center. By making use of the center-based pipeline and carefully designed constraints, our approach achieves the very promising performance in both weakly-supervised and fully-supervised settings. As shown in Table~\ref{table2}, our weakly-supervised model outperforms all other model-based methods while our fully-supervised method further improves the performance. We show the visual results of our method and the current state-of-the-art weakly-supervised method S2HAND~\citep{chen2021s2hand} in Fig.~\ref{Visualcompare}. It can be seen that our method has obvious advantages in both texture detail and pose estimation.
%As a universal multi-hand framework, our model does not require camera intrinsics and ground truth scale. 

% \begin{figure}[!h]
%     \centering
%     \includegraphics[width=0.94\columnwidth]{images/RHD-mesh-less.png}
%     \vspace{-0.14in}% Reduce the figure size so that it is slightly narrower than the column.
%     \caption{Predictions results under two-hand setting. The first two rows are the results from RHD and the last two rows are results from InterHand.}
%     \label{RHDmesh}
%     \vspace{-0.2in}
% \end{figure}

\begin{table}[!ht]
\caption{Performance comparison with multi-stage methods on multi-hand datasets. We compare the results of our multi-hand model and S2HAND~\citep{chen2021s2hand} equipped with a third-party detector~\citep{Shan2020UnderstandingHH}, and the results of S2HAND directly using ground-truth bounding boxes, evaluating 2D keypoint distances (pixels) and 3D joint point errors (cm).}
\centering
\resizebox{.95\columnwidth}{!}{
\begin{tabular}{c|ccc}
\toprule
    Methods & S2HAND +Detector & Ours\_Multi & S2HAND+GT   \\
\midrule
    2D Distance$\downarrow$ & 9.79 &  \textbf{7.41} &  9.54   \\
\midrule    
    MPJPE$\downarrow$ & 1.55& \textbf{0.95}& 1.52\\
\bottomrule
\end{tabular}}

\label{MultiCompare}
\end{table}

\begin{figure}[ht]
    \centering
    \includegraphics[width=0.96\columnwidth]{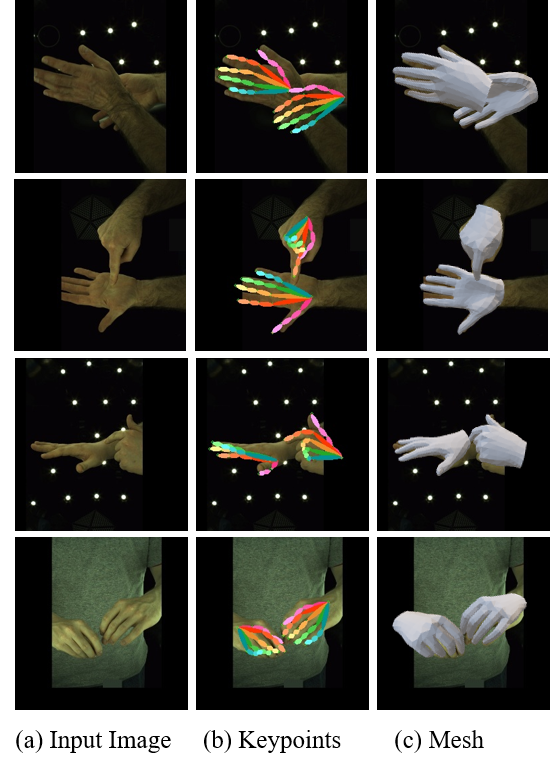} 
    \vspace{-0.14in}% Reduce the figure size so that it is slightly narrower than the column.
    \caption{Qualitative 3D pose estimation results on InterHand2.6M dataset.}
    \label{visualinterac}
    \vspace{-0.2in}
\end{figure}

% \begin{table}[!ht]
% \caption{Mesh reconstruction with different decoders (MANO and GCN from IntagHand) on the FreiHAND testset trained on a reduced trainingset using 3D joints only.}
% \centering
% \resizebox{.95\columnwidth}{!}{
% \begin{tabular}{c|ccc}
% \toprule
%     Decoder Type & MPJPE$\downarrow$ & MPVPE $\downarrow$ & training time   \\
% \midrule
%     MANO & 0.879 &  0.886 &  1.5 days   \\
% \midrule    
%     GCN & 0.776 & 0.792 & 5 days \\
% \bottomrule
% \end{tabular}}

% \label{MultiCompare}
% \end{table}

\noindent \textbf{Two-hand Reconstruction Experiments} Secondly, we evaluate our method on RHD and InterHand2.6M with two isolated hands and interacting hand scenarios, respectively. 

As for RHD, we compare our method with several fully-supervised methods~\citep{zimmermann2017learning,yang2019disentangling,spurr2018cross,moon2020interhand2}, where the EPE results in Table~\ref{table3} are reported from~\citep{moon2020interhand2}. All of the above methods only predict the sparse hand joints rather than dense hand mesh, they require the ground truth bounding box to crop the hand areas. To facilitate the fair comparisons, we train our model with the same cropped images and evaluate the relative 3D joints error. It can be seen that our weakly-supervised model achieves the promising result without requiring 3D supervision, ground truth scale or hand type. We obtain 2.07 cm end point error for 3D joints and 8.09 pixel error under input resolution of $224\times 224$. Differently from single-hand setting, it is challenging to distinguish between left and right hand types while recovering 3D pose. We achieve 97.65\% accuracy for hand type classification. As a single stage pipeline, we can detect and recover hands, simultaneously. Furthermore, we train our model using the original image, which achieve the 2.10 cm end point error for 3D joints and 9.14 pixel error under input resolution of $320\times 320$. The tiny accuracy loss demonstrates the effectiveness of our proposed single-stage pipeline. 
% Fig.~\ref{RHDmesh} shows the visual results on RHD evaluation set, including input image, the predicted mask, keypoints and mesh overlaid with input image.

To examine the performance of our presented method on images captured in the real world, we select 200K images from the training set of InterHand2.6M to train the model and use the whole testing set for evaluation. It spends lots of computational cost on training all the data together. Similar to RHD, we firstly train our model with the cropped images and evaluate the relative 3D joints error. We achieve 2.77 cm end point error and 10.98pixel error under the input size of $224\times 224$. Then, we train our model using the original image without cropping. It achieves 2.39 cm end point error for 3D joints and 15.82 pixel error under input resolution of $512\times 512$. Due to the difference between our fixed focal length in training and the ground truth, the predicted 3D coordinates and the true value from the dataset cannot be completely matched. Therefore, the 2D loss in this experiment can more accurately reflect the performance of our presented method. Visualization results are shown in Fig.~\ref{visualinterac}.
% Fig.~\ref{RHDmesh} shows the visual results on InterHand2.6M. Although our method is not specifically designed for interacting scenarios and each hand is detected and reconstructed independently, our model is able to recover interacting hands heavily occluded from each other. 
In further work, we consider processing a group of two interacting hands together like InterHand~\citep{moon2020interhand2} or introduce other assumptions such as collision detection and left-right hand association to improve the accuracy.

\begin{table}[!ht]
\caption{Comparison on different combinations of loss terms tested on the evaluation set of FreiHAND. Each loss term can improve the performance of the model to a certain extent, and the photometric loss mainly predicts texture and lighting to restore realistic meshes.}
\centering
\resizebox{1.0\columnwidth}{!}{
\begin{threeparttable}
\begin{tabular}{ccccccccc}
\toprule
    Baseline & $\mathcal{L}_{reg}$ & $\mathcal{L}_{aug}$ & $\mathcal{L}_{con}$ & $\mathcal{L}_{bone}$ &$\mathcal{L}_{pho}$ & $\mathcal{L}_{3D}$ & MPJPE$\downarrow$ & MPVPE$\downarrow$ \\
\midrule
    $\checkmark$ & - & - & - & - & - & - & 2.98 & 3.16  \\
    $\checkmark$ & $\checkmark$ & - & - & - & - & - & 1.66 & 1.72  \\
    $\checkmark$ & $\checkmark$ & $\checkmark$ & - & -  & - & - & 1.53 & 1.61  \\
    $\checkmark$ & $\checkmark$ & - & $\checkmark$  & - & - & - & 1.64 & 1.71  \\  
    $\checkmark$ & $\checkmark$ & -& $\checkmark$ & $\checkmark$  & - & - & 1.43 & 1.45  \\
    % $\checkmark$ & $\checkmark$ & $\checkmark$ & $\checkmark$ & $\checkmark$norot & - & - & 1.14 & 1.17  \\
    % $\checkmark$ & $\checkmark$ & $\checkmark$ & $\checkmark$ & $\checkmark$norot & - & $\checkmark$ & 0.99 & 0.98  \\    
    $\checkmark$ & $\checkmark$ & $\checkmark$ & $\checkmark$ & -  & - & - & 1.48 & 1.56  \\  
    $\checkmark$ & $\checkmark$ & $\checkmark$ & - & $\checkmark$  & - & - & 1.13 & 1.17  \\     
    $\checkmark$ & $\checkmark$ &$\checkmark$ & $\checkmark$ & $\checkmark$ &  - & - & 1.07 & 1.10  \\    
    $\checkmark$ & $\checkmark$ & $\checkmark$ & $\checkmark$ & $\checkmark$ & $\checkmark$ & - & \textbf{1.07} & \textbf{1.10}\tnote{*}  \\    
    $\checkmark$ & $\checkmark$ & $\checkmark$ & $\checkmark$ & $\checkmark$ & - & $\checkmark$ & 0.80 & 0.81  \\     
    $\checkmark$ & $\checkmark$ & $\checkmark$ & $\checkmark$ & $\checkmark$ & $\checkmark$ & $\checkmark$ & \textbf{0.80} & \textbf{0.81}\tnote{*}  \\     
\bottomrule
\end{tabular}
 \begin{tablenotes}
        \footnotesize
        \item[*] this is the dense mesh with vivid hand texture. 
      \end{tablenotes}
\end{threeparttable}}
\label{ablation}
\vspace{-0.12in}
\end{table}

\begin{table}[!ht]
\caption{PSNR comparison of reconstruction results tested on evaluation set of FreiHAND and HO3D. Ours\_mean denotes our model using mean texture.}
\centering
%\resizebox{.95\columnwidth}{!}{
\begin{tabular}{ccc}
\toprule
    Dataset & FreiHAND$\uparrow$ & HO3D$\uparrow$ \\
\midrule
    S2HAND~\citep{chen2021s2hand} & 14.74 & 13.92  \\
    Ours\_mean & 11.79 & 13.71 \\
    Ours & \textbf{16.64} & \textbf{16.78} \\
\bottomrule
\end{tabular}

\label{PSNR}
\vspace{-0.1in}
\end{table}

\begin{figure}[ht]
    \centering
    \includegraphics[width=0.98\columnwidth]{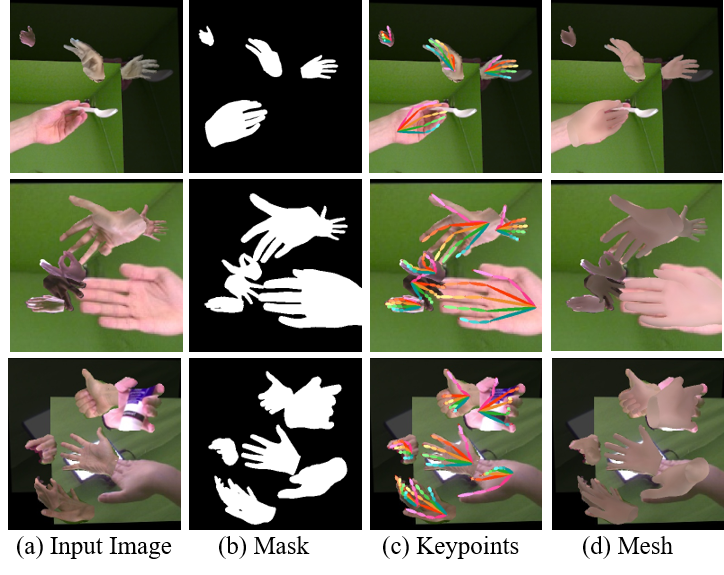} 
    \vspace{-0.14in}% Reduce the figure size so that it is slightly narrower than the column.
    \caption{Qualitative 3D pose estimation results on the proposed multi-hand dataset. From left to right: generated image, predicted mask, predicted keypoints and predicted mesh overlaid on input image.}
    \label{visualmulti}
    \vspace{-0.2in}
\end{figure}

\begin{figure}[ht]
    \centering
    \includegraphics[width=0.98\columnwidth]{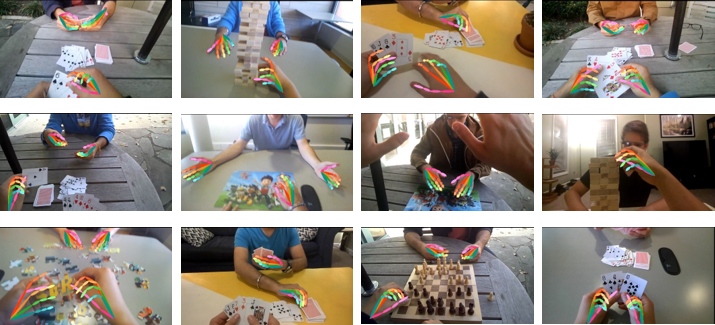} % Reduce the figure size so that it is slightly narrower than the column.
    \caption{Qualitative 3D pose estimation results on images in the wild. The predicted keypoints are re-projected from estimated 3D mesh using our model, which is trained only on our generated multi-hand data.}
    \label{egohand}
    \vspace{-0.2in}
\end{figure}

\noindent \textbf{Multi-hand Reconstruction Experiments} To verify the effectiveness of our method in multi-handed scenarios, we use the data augmentation strategy proposed above to generate a new validation dataset for experiments. ~Since there are few off-the-shelf methods that can handle the multi-hand reconstruction, we solve it by equipping the existing single-hand reconstruction method with an extra detector. We choose S2HAND~\citep{chen2021s2hand} for comparison, since it adopts a similar model-based approach and provides training code as well as evaluating models. We employ EgoHand~\citep{Shan2020UnderstandingHH} as the detector. It is specially trained from an egocentric perspective that is close to the multi-hand data we want to test on. 
% The training and evaluation data are created based on the publicly available single hand datasets through our weakly-supervised data augmentation scheme. In order to ensure the clear visibility of each hand, we limit up to ten visible hands during training. 
We compare the 2D pixel error and MPJPE of our multi-hand model with the detector-equipped S2HAND. To examine the influence of the detector on the reconstruction accuracy, we test the results of directly utilizing the ground-truth hand bounding box as input. As shown in Table~\ref{MultiCompare}, our method achieves clear advantages in both cases. Fig.~\ref{visualmulti} depicts the visual results of our multi-hand model, including input image, the predicted mask, keypoints and mesh overlaid with the input image, respectively. 
% Besides, we regress the lighting and texture parameters to obtain the more vivid hand mesh. By taking advantage of our assumption on uniform camera intrinsic, we can estimate the absolute hand position and orientation without the complicated transformations.
% Compared to the multi-stage methods, our model only needs single forward inference, which avoids the redundant feature encoding for each hand. 
In order to investigate the generalization ability of our proposed approach, we evaluate our model on the unlabeled images from Bambach et al.~\citep{Bambach_2015_ICCV}. As shown in Fig.~\ref{egohand}, our method can obtain the reasonable prediction results even without fine-tuning. To achieve better accuracy in real scenes, the model can be fine-tuned using manually annotated 2D keypoints in the target scene, since 2D labels are easier to obtain than 3D labels.

\subsection{Ablation Study}
\begin{figure}[ht]
    \centering
    \includegraphics[width=0.94\columnwidth]{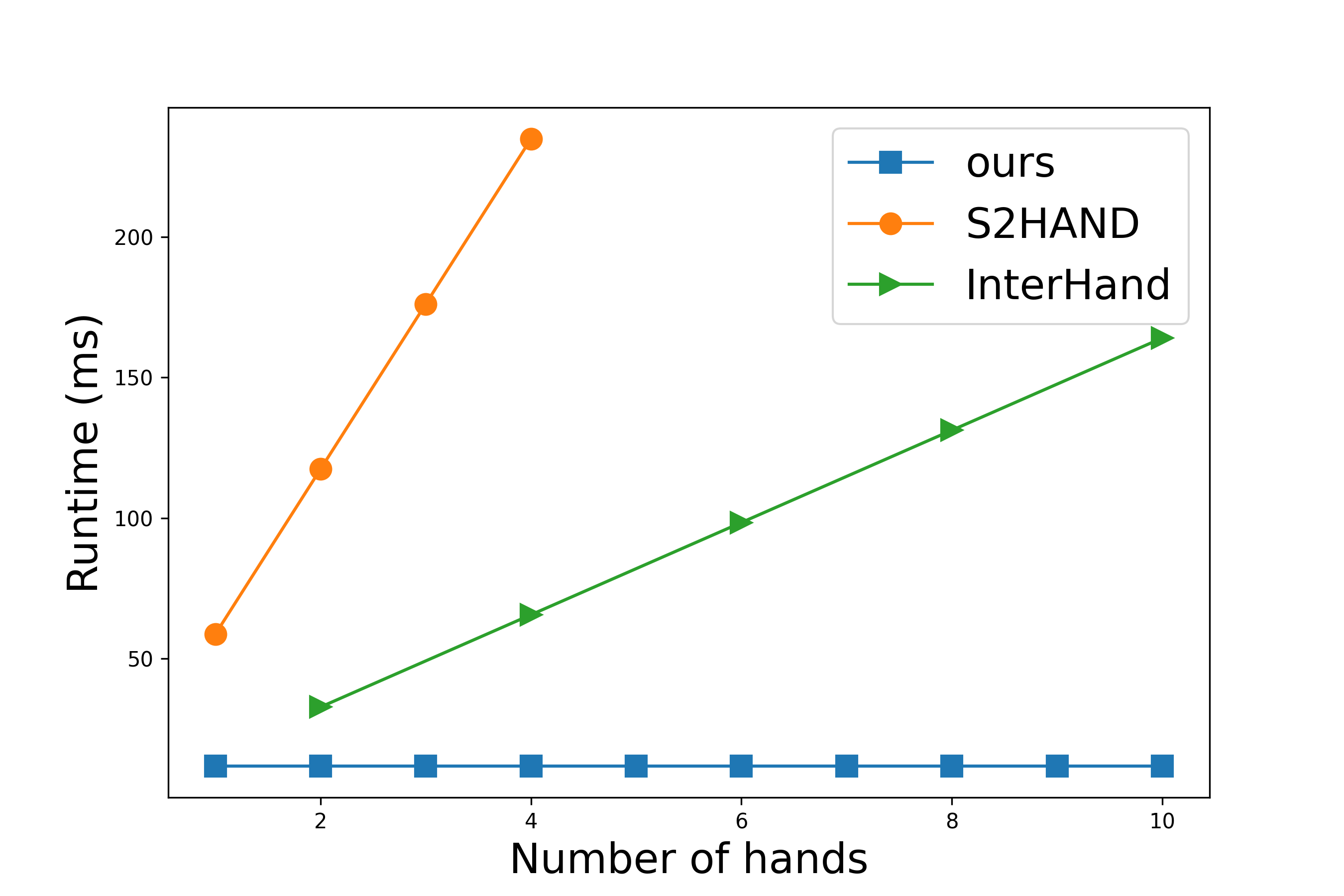}
    \vspace{-0.12in}% Reduce the figure size so that it is slightly narrower than the column.
    \caption{Runtime comparison on the different number of hands. The computational cost grows linearly using a multi-stage pipeline, while our single-stage pipeline only needs a single forward pass. }
    \label{runtime}
\end{figure}

\noindent \textbf{Evaluation on Efficiency}
The conventional methods using multi-stage pipeline need to detect and encode each hand patch individually while our presented network shares the same feature map only requiring single forward pass for inference. For the single-hand setting, we employ the input image with the size of $224\times 224$. To facilitate the fair comparison, we conduct the experiments on the same device, and use the official implementation of the reference methods. Our model only takes 11.8ms for inference, while S2HAND~\citep{chen2021s2hand} spends 58.7ms and InterHand~\citep{moon2020interhand2} requires 16.4ms with the same input. It can be seen that our model is the most lightweight under the same conditions. As for the multi-hand setting, the computation cost of multi-stage methods grows linearly with the number of hands in image, as depicted in Fig.~\ref{runtime}. In addition, detection and cropping time need to be considered, which incurs the extra computation cost and requires off-the-shelf detectors. Besides, we find that the running time of our model mainly depends on the size of input image. The inference time with the size of $512\times 512$ is 36.5ms, which is still faster than S2HAND. Through this experiment, we believe that the single-stage framework we proposed has its merit in dealing with multiple hands.

\noindent \textbf{Hand Center}
We study several center definitions such as the center of bounding box, mean position of visible landmarks and fixed joint position like wrist or first part of middle finger. Firstly, the fixed joint position is unsuitable for all kinds of datasets. In some datasets, the invisible joints are set to -1, which makes this definition not applicable when the fixed center location is invisible. Then, we test the accuracy of center definition on FreiHAND using the center of bounding box and mean position of visible landmarks. The former achieves 1.09 cm MPJPE and 1.12 cm MPVPE while the latter obtains 1.07 cm MPJPE and 1.10 cm MPVPE. In some poses, the center of the bounding box may fall on background pixels outside the hand, while the center of the mean position of visible landmarks can mostly fall on the area belonging to the hand. Therefore, we choose the latter for its robustness.

\begin{figure}[ht]
\centering
 \includegraphics[width=0.48\textwidth]{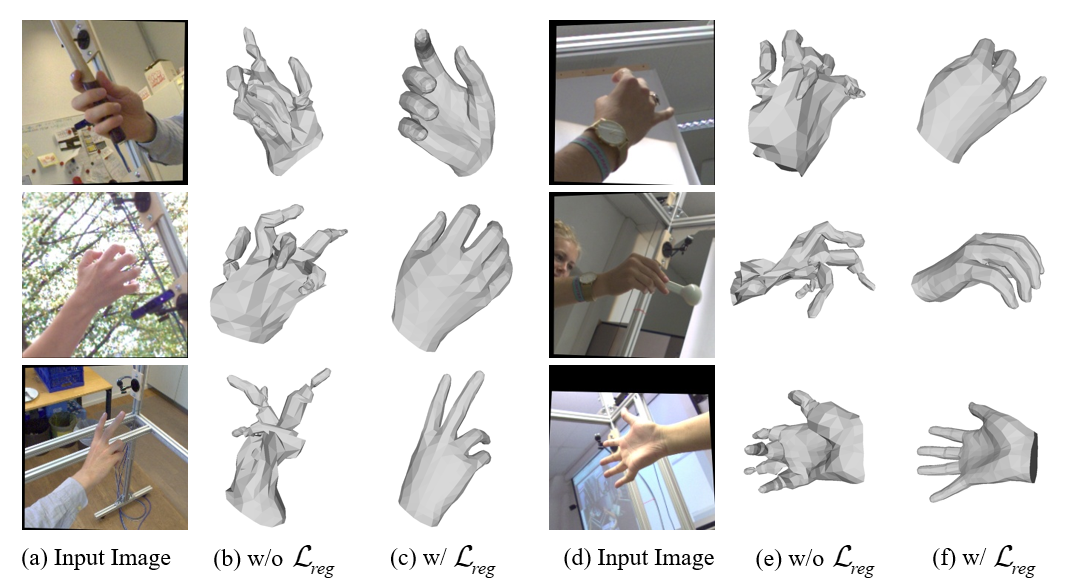} 
 \vspace{-0.12in}% Reduce the figure size so that it is slightly narrower than the column. 
\caption{Visual comparison of model trained with and without pose regularization term $\mathcal{L}_{reg}$. Models without pose regularization constraints may generate implausible hand poses.}
\label{regularization}
\vspace{-0.2in}
\end{figure}

\noindent \textbf{Effect of Different Loss Terms}
Finally, we conduct a comprehensive comparison on different loss terms. The overall comparison results on FreiHAND dataset are depicted in Table~\ref{ablation}. The re-projected keypoints error is the most fundamental loss function for our weakly-supervised pipeline, which is treated as a baseline. $\mathcal{L}_{bone}$ is the second term in $\mathcal{L}_{rep}$ that introduces constraint on 2D bone direction. It provides more detailed pose information, which plays an import role in our weakly-supervised algorithm. $\mathcal{L}_{cons}$ introduces the top-down and bottom-up consistency, which further improves the overall accuracy. $\mathcal{L}_{pho}$ does little improvement for pose accuracy, since other losses have been able to constrain the optimization direction of the model. However, the results without $\mathcal{L}_{pho}$ are with purely gray texture. Compared to S2HAND, our method achieves higher PSNR (Peak Signal to Noise Ratio) scores, which means our predictions are more realistic, as shown in Fig.~\ref{Visualcompare}. Detailed PSNR results can be found in Table~\ref{PSNR}. $\mathcal{L}_{reg}$ is adopted to avoid the implausible 3D poses, which makes the limited contribution to the final accuracy. In some cases, it even reduces the accuracy. However, a lower loss with the unreasonable hand shape is not the expected result, which often means overfitting. The difference between with and without $\mathcal{L}_{reg}$ is depicted in Fig.~\ref{regularization}. Besides, $\mathcal{L}_{aug}$ is not a specific loss term, which refers to whether to use the data augmentation strategy mentioned above during training. The data augmentation can significantly improve the model accuracy, which can avoid overfitting and make full use of the underlying pose distribution. $\mathcal{L}_{3D}$ refers to the extra joint constraints when 3D supervision exists. The loss term is the same as $\mathcal{L}_{cons}$ except that the constrained objects are 3D joints.

% Other novel loss functions introduced by previous research works may significantly improve the final accuracy, such as the 2D-3D consistence loss introduced in~\citep{chen2021s2hand} and biomechanical constraints presented in~\citep{spurr2020weakly}. However, in order to verify the effectiveness of our proposed framework, we only kept the necessary loss function. 

\begin{figure}[ht]
\centering
\includegraphics[width=0.4\textwidth]{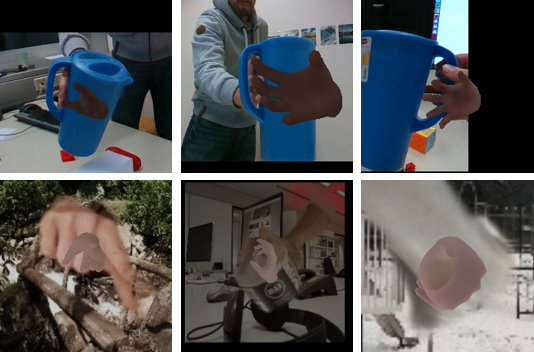}
\vspace{-0.12in}% Reduce the figure size so that it is slightly narrower than the column.
\caption{Failure cases, including extreme hand pose and texture.}
\label{failcase}
\vspace{-0.14in}
\end{figure}

\noindent \textbf{Limitations}
With only 2D supervision, it is difficult for our model to handle the ambiguity of some specific gestures. Specifically, two gestures that are symmetrical with respect to the projection plane are identical in the 2D projection view. Additionally, it is difficult for our model to get the accurate result when the input gesture is too challenging. Fig.~\ref{failcase} shows some failure cases, including object occlusion, motion blur, extreme texture and lighting.

\section{Conclusion}
This paper proposed a novel approach to simultaneously locating and recovering multiple hands from single 2D images. In contrast to the conventional methods, we presented a concise but efficient single-stage pipeline that reduced the computational redundancy in data preprocessing and feature extraction. Specifically, we designed a multi-head auto-encoder structure for multi-hand recovery, where each head network shares the same feature map and outputs hand center, pose and texture, respectively. Besides, a weakly-supervised scheme was proposed to alleviate the burden of expensive 3D real-world data annotations. Extensive experiments on the benchmark datasets demonstrate the efficacy of our proposed framework. Our method achieved the promising results comparing to the previous state-of-the-art model-based methods in both weakly-supervised and fully-supervised settings. In further work, we intend to extend our work to real-time hand gesture recognition~\citep{Lazarou2021ANS} and human-scene interaction~\citep{Hassan2021Populating3S} using multi-view contrastive learning and temporal consistency. It may reduce the burden of 3D annotation and achieve the high reconstruction accuracy.

\bibliographystyle{model2-names}
\bibliography{refs}

\end{document}